\def\year{2019}\relax
\documentclass[letterpaper]{article} %DO NOT CHANGE THIS
\usepackage{aaai19}  %Required
\usepackage{times}  %Required
\usepackage{helvet}  %Required
\usepackage{courier}  %Required
\usepackage{url}  %Required
\usepackage{graphicx}  %Required
\usepackage{latexsym}
\usepackage{amsfonts,amssymb}
\usepackage{amsmath}
\usepackage{enumitem}
\usepackage{graphicx}
\usepackage{subfigure}
 \usepackage{mathrsfs}
 \usepackage{multirow}
 \usepackage{booktabs}
 \usepackage{algorithm}
 \usepackage{algorithmic}
 \usepackage{array}
\graphicspath{{Figures/}}
\begin{document}
	\title{SEGAN: Structure-Enhanced Generative Adversarial Network for\\ Compressed Sensing MRI Reconstruction}
	\author{Zhongnian Li, Tao Zhang, Peng Wan, Daoqiang Zhang\thanks{This research was supported by the National Natural Science Foundation of China ( 61876082, 61861130366, 61703301, 61473149 ).}\\
	College of Computer Science and Technology,\\ Nanjing University of Aeronautics and Astronautics, Nanjing, China \\
	\{zhongnianli,dqzhang\}@nuaa.edu.cn	}
	\maketitle
	\begin{abstract}
		Generative Adversarial Networks (GANs) are powerful tools for reconstructing Compressed Sensing Magnetic Resonance Imaging (CS-MRI). However most recent works lack exploration of structure information of MRI images that is crucial for clinical diagnosis. To tackle this problem, we propose the Structure-Enhanced GAN (SEGAN) that aims at restoring structure information at both local and global scale. SEGAN defines a new structure regularization called Patch Correlation Regularization (PCR) which allows for efficient extraction of structure information. In addition, to further enhance the ability to uncover structure information, we propose a novel generator SU-Net by incorporating multiple-scale convolution filters into each layer. Besides, we theoretically analyze the convergence of stochastic factors contained in training process. Experimental results show that SEGAN is able to learn target structure information and achieves state-of-the-art performance for CS-MRI reconstruction.
	\end{abstract}

	\section{Introduction}
	Magnetic Resonance Imaging (MRI) is a promising technique for disease diagnosis. The biggest advantage of MRI is no risk of radiation that is inevitable in other medical imaging techniques, such as Computed Tomography (CT). However, for patients, lengthy scan process still causes discomfort unavoidably \cite{hollingsworth2015reducing}. To shorten MRI acquisition time, Compressed Sensing (CS) \cite{candes2006robust,candes2008introduction} has been applied to MRI sampling and reconstruction \cite{lustig2007sparse}. With the assumption of signals sparsity, CS reconstructs MRI images from under-sampled signals. The relationship between under-sampled MRI and full-sampled MRI can be linearly formulated as follows,
	\begin{equation}
	\ y = Ax + \xi  
	\end{equation}
	\begin{figure}
		
		\includegraphics[width=0.46\textwidth]{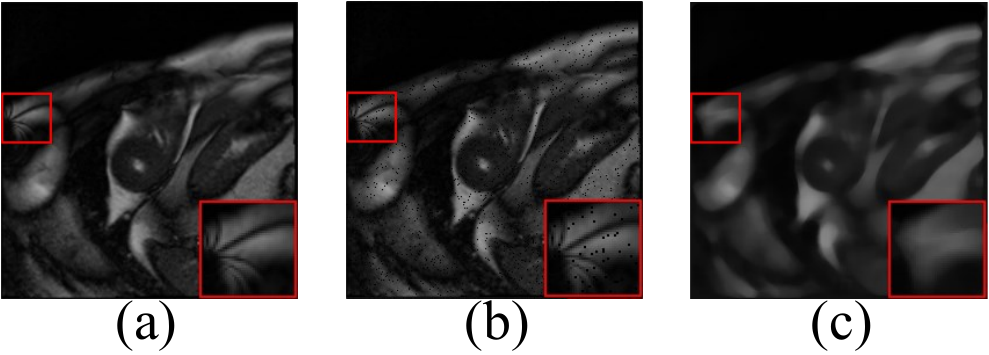}
		\centering	
		\caption{Comparision of MRI images with different types of structure similarity, all with Mean Squared Error (MSE)= 73. (a) Original image. (b) salt-pepper impulsive noise contaminated image. (c) Blurred image. Although two images have the same MSE value, the structure similarity of third image is lower than second image. Human vision system might feel more comfortable with second image in medical diagnosis.}
		\centering
		\label{structure}
	\end{figure}where the vector $y\in$ $\mathbb{C}^{m}$ is  observed vector and  $x\in$ $\mathbb{C}^{N}$ is raw data. Matrix $A$ is the under-sampled Fourier encoding matrix and $\xi$ is noise vector. The objective of compressed sensing MRI (CS-MRI) is to recover $ x $  from $ y $. The challenge arises from the under-sampled cases $ m\ll N $. Clearly, reconstructing task is essentially under-determined and priori knowledge is absolutely necessary to obtain a unique solution. Among those priori knowledges, sparsity is the most effective and widely applied characteristic \cite{goldstein2009split,lai2016image} and the performance of reconstruction is guaranteed in theory \cite{candes2008introduction}.

	Although sparsity assumption achieves many successes in various applications, the hypothesis still has a great deal of limitations (e.g., structure sparse, etc.) \cite{jia2012visual}. To address the problems resulting from the sparsity assumption, deep learning based models are introduced for CS-MRI reconstruction which only requires pairwise under-sampled images and the corresponding full-sampled ones \cite{shen2017deep,SchlemperCHPR17,HammernikKKRSPK17}. Generally, the objective function of traditional deep network based model is composed of Minimal Squared Error (MSE) and certain kinds of regularizations. For Generative Adversarial Networks (GANs) based models, cost function is built on the divergence between the distribution of generated images and that of real ones\cite{goodfellow2014generative,Yang2017DAGAN,Quan2018Compressed}. 
	
	Howerver, both kinds of loss functions can not explicitly quantify structure correlation. Note that, for MRI reconstruction task, structure information consists in correlations between any two patches of an image. Correlation can be reflected in different aspects, such as the similarity between two patches. It is obvious that structure information plays a key role in clinical diagnosis, such as pathologic analysis, personalized treatment. Hence, it is of great importance for doctors to access reconstructed images that preserve structure information well. To gain a more intuitive understanding of structure information, we take images with same MSE for example, as illustrated in Figure \ref{structure}, the amount of valuable structure information contained in the three images could have a great difference.
	
	In conclusion, explicitly exploring the restoration of structure information is indispensable in CS-MRI, which inspires us to propose a new deep generative model SEGAN to restore structure information contained in original MRI images. Three main contributions of our work are summarized as follows:

\begin{enumerate}[label=\arabic*)]
	\item We propose Patch Correlation Regularization (PCR) to capture the local structure information of MRI images. Further, we introduce the Stochasic PCR in order to accelerate the model training without performance degradation. Besides, the measure of global structure similarity  incorporated into loss function as well for preserving the   whole structure to some extent.
	\item We propose a novel generator called SU-Net that comprises different sizes of convolution filters at each layer for facilitating the capture of multiscale structure information.
	\item 
	We present a theoretical analysis of the convergence of our approach using the online programming framework, which is optimized by gradient descent.
\end{enumerate}

	\section{Related work}
	In this paper, compressed sensing and deep generative model are combined for MRI images reconstruction. Hence, we discuss the related work in two portions. The first portion briefly introduces conventional works on CS-MRI and the second portion reviews recent researches of MRI reconstruction based on the deep generative model.
	
	Conventional CS-MRI reconstruction methods assume that sparsity of signal is a natural characteristic, which has been fully utilized for recovering full-sampled signal. However, computation cost of Sparsity-based methods of reconstruction is huge \cite{goldstein2009split} where complex nonlinear  minimization problem needs to be optimized. In recent years, low rank \cite{yao2015accelerated,otazo2015low} is another assumption for reconstruction which views CS-MRI as low-rank completion and requires to solve computationally expensive nuclear norm as well. To break the limitations of transform-based method, dictionary learning\cite{aharon2006k,caballero2014dictionary,zhan2016fast} attempts to learn data-dependent dictionary rather than bounded by existing signal transformation methods, but redundant atoms make time complexity unacceptable in real-world applications. 
	
	The deep generative model is one of the most active research topics in CS-MRI reconstruction. One line of work attempts to reconstruct CS-MRI by virtue of deep learning. Directly mapping zero-filling reconstruction on full-reconstruction with convolutional autoencoder network has been explored in the early works \cite{Lee2017Deep}. Deep ADMM-Net \cite{Yang2017ADMM} is anther approach to solve CS-MRI of which deep architecture is derived from the ADMM optimization algorithm. Recently, GANs, a great promising model in image generation, are introduced to reconstruct MRI images as well. Cyclic data consistency loss has been employed for improving the performance of CS-MRI reconstruction with GANs \cite{Quan2018Compressed}. DAGAN \cite{Yang2017DAGAN} incorporates content loss composed of pixel-wise Mean Square Error (MSE) and Frequency Error (FE) into objective function to enhance the performance of generator. 
	In conclusion, the above discussed works have not tackle the restoration of structure information without exception.

	\section{The Proposed Method}
		\subsection{Problem Statements}
	 The main reconstruction task of compressed sensing MRI (CS-MRI) is to recover raw data space  $\mathcal{X} $ from under-sampled data space  $\mathcal{Y} $ and the biggest challenge is that the under-sampled rate is much less than Shannon sampling rate. Let training set be $ S = \{ ({x^1},{y^1}),({x^2},{y^2}),...,({x^l},{y^l})\}  $ , we denote $X = {[{x^1},...,{x^l}]^T}\in \mathbb{R}^{l \times N}  $ and $Y = {[{y^1},...,{y^l}]^T}\in \mathbb{R}^{l \times m} $ , where $ x^{i} $ is $ i^{th} $ original image and $ y^{i} $ is the $ i^{th} $ under-sampled image. 
	 \subsection{The Formulation}	 
	In this paper, we propose Structure-Enhanced Generative Adversarial Network (SEGAN) for generating structurally informative CS-MRI images. The objective function of SEGAN is composed of four parts, including pixel loss, Patch Correlation Regularization (PCR), Measure Structure Similarity Regularization (SSIMR) and Generative Adversarial Loss.  The loss function can be written as follows:
\begin{equation}
\begin{aligned}
	\min _G{\max _D}\:L_{se}(D,G)\: & ={\mathbb{E}_{x \sim P_{data}(x)}}[\log D(x)] \\ &+ {\mathbb{E}_{y\sim P_{data}'(y)}}[1 - \log D(G(y))] \\ &+ \sum\limits_{i = 1}^n ({{\lambda _1}{P_{{w_G}}}({x^i},G{{(y^i)}})}  \\ &+ {\lambda _2}{S_{{w_G}}}({x^i},G{(y^i)}) \\ & + {\lambda _3}{\left\| {{x^i},G{{(y^i)}}} \right\|_F})
	\end{aligned}
\end{equation}
     $ G(y^{i}) $ denotes the  $ i^{th} $ generated image. ${{S_{{w_G}}}({x^i},G{{(y^i)}})} $ is SSIMR for restoring global structure information, $ {P_{{w_G}}}({x^i},G{(y^i)}) $ is PCR for extracting local structure information. $ P_{data}(x) $ denotes the distribution of full-sampled image and $ P_{data}'(y) $ denotes the distribution of under-sampled  MRI image.
     \begin{figure*}[t]
     	\centering
     	\includegraphics[width=\textwidth]{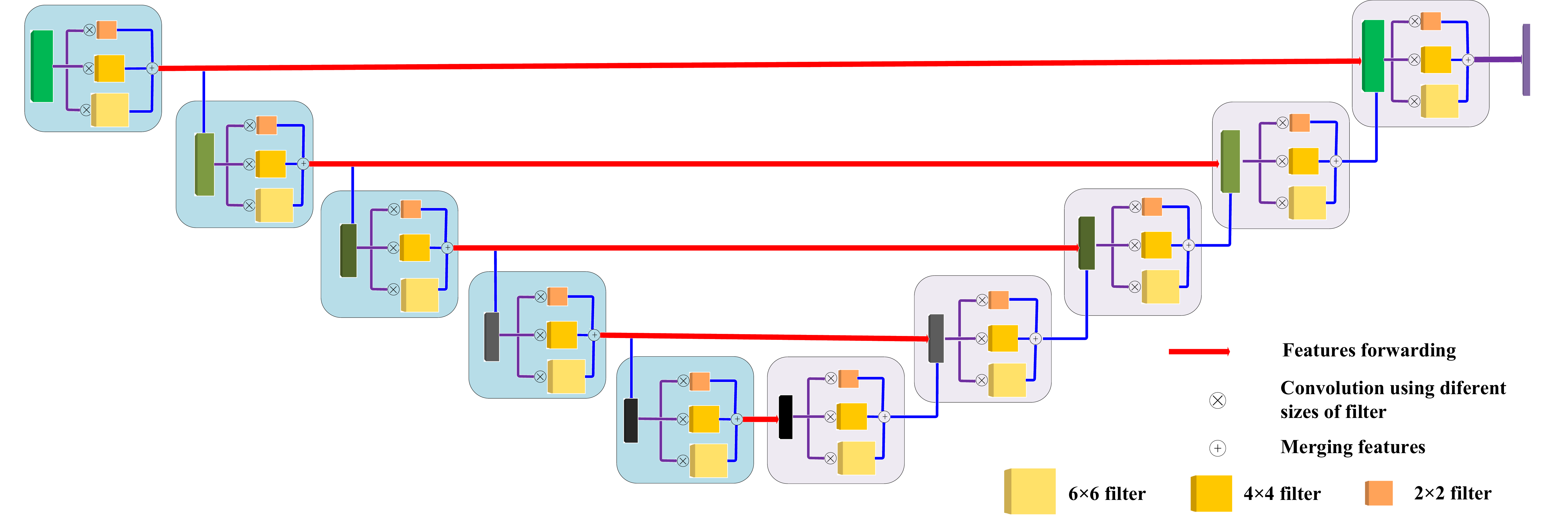}
     	\centering
     	\caption{Illustration of the proposed SU-Net using multi-scale convolutional filters. Each unit is composed of three main components: 1) feature forward way, 2) three sizes of convolutional filters, 3) leaky ReLU. In our designed SU-Net, there are 10 such units that fully exploit structure information contained in multiple patches.}
     	\label{u-net}
     \end{figure*}
    \subsection{Structure-Enhanced Loss}
    Overall, the loss of generator, named as SEL, consists of three parts: PCR, SSIMR and Mean Squared Error (MSE), SEL can be written as,
\begin{equation}
\begin{aligned}
    \ L _{{w_G}} = &\sum\limits_{i = 1}^n ({{\lambda _1}{P_{{w_G}}}({x^i},G{{(y^i)}})}  + {\lambda _2}{S_{{w_G}}}({x^i},G{(y^i)}) \\&  + {\lambda _3}{\left\| {x^i,G(y^i)} \right\|_F})
    \end{aligned}
    \end{equation}
	where $\lambda_1, \lambda_2, \lambda_3$ are trade-off parameters that control the importance of the three regularization respectively. $\left\| \:\cdot\: \right\|_F$ denotes Frobenius norm.
	
	\subsubsection{Patch Correlation Regularization}
	
	According to prior knowledge, MRI image belongs to structure data in which there is strong correlation among the pixels in the neighborhood \cite{tremoulheac2014dynamic}. Among the existing methods, graphs based models are the most common methods for processing structure data, which inspires us to design a graph based structure regularization called Patch Correlation Regularization (PCR) that is dedicated to preserve intrinsic local correlation, where weights of the graph are in proportion to correlation of two image patches.
	
	We split single MRI image into N equal size patches and calculate the correlation between patches. The same operation also acts upon outputs of generator.  $ G(y')_i $ is $ i^{th} $ patch of the output of generator and $ x_i' $ is $ i^{th} $ patch of MRI image. PCR can be described as follows:
\begin{equation}
\begin{split}
\ P_{{w_G}}(x',G(y)') = \sum\limits_{i = 1}^N \sum\limits_{j = i + 1}^N [f(x_i',{x_j'})\\ - f(G(y')_i,G(y')_j)]^2 
\end{split}
\end{equation}
	Where $ f(x_i',x_j') $ is function of computing matrix correlation.

	 \subsubsection{Correlation Function Selection}For structure data, the type of measured structure information is determined by the selection of correlation function. Generally, correlation functions can be divided into three categories: kernel function, distance metric and correlation coefficients.  Here, we take Pearson correlation coefficient and Gauss kernel for example, the former is widely used to obtain linear measure of correlation in MRI image, the latter is more common in obtaining nonlinear measure.
	 
	 Given a differentiable correlation function set $ M = \{ {f_1}(x,y),{f_2}(x,y), \ldots ,{f_m}(x,y)\} $ with corresponding weights vector   $ u = {[{u_1},{u_2}, \ldots ,{u_m}]^T} \in { \mathbb{R}^m} $ , so we have various linear combinations of correlation functions $ F = \{ f(x,y)\:|\: f(x,y) = \sum\limits_{i = 1}^m {u_i{f_i}(x,y)} ;x,y \in \ \mathbb{R}^{n\times n}\}  $ . $ u $ is selected by domain knowledge.

	\subsubsection{Stochastic Patch Correlation Regularization}
	 In practice, PCR comes at the cost of large accessing memory burden, so we propose a method called Stochastic PCR for accelerating training, it can be formulated as follows:
\begin{equation}
\begin{split} 
\ P_{{w_G}}(x',\:G(y')) = \sum\limits_{i = 1}^N \sum\limits_{j = i + 1}^N {\alpha _{i,j}}[f(x_i',\:x_j')\\ - f(G(y')_i,\:G(y')_j)]^2
\end{split}
\end{equation}

$\alpha _{i,j} $ is a random variable obeying the $0-1$ distribution. Note that, larger memory of GPU allows for more patches selected for training, so mean of $\alpha _{i,j} $ is determined by hardware capacity. Hence, we have a randomized version of Structure-Enhanced Loss called SSEL.
    	\begin{algorithm}[htb] 
    	\caption{ Structure-Enhanced GAN} 
    	\label{alg:Framwork} 
    	\begin{algorithmic} %这个1 表示每一行都显示数字
    		{\bf \STATE Input:\\ }  
    		$\eta$ Learning rate, $ n $ number of iterations for training the discriminator, N patch size, M batch size\\
    		{\bf \STATE Output:\\ } 
    		Structure-Enhanced GAN generator parameters $w_G$
    		{\bf \STATE Initialize }
 			 $w_G$, $w_D$
    		\REPEAT 
    		\FOR{$i=1$ to $n$ } 
    		\STATE Sample a minibatch $  x^i, x^i \sim P_{data}(x)$.
    		\STATE Sample a minibatch $ y^i, y^i\sim P_{data}'(y) $.
    		\STATE Calculate $ G(y^i)$
    		\STATE $g_{w_D}\leftarrow{\nabla _{{w_D}}}(\frac{1}{M}\sum\limits_{i = 1}^M \log D({x^i})  $
    		\STATE$\qquad\quad  + \log (1 - D(G'({y^i})))) $
    		\STATE ${w_D} \leftarrow {w_D} + \eta $ \bf ADAM$(w_D,g_{w_D})$
    		\ENDFOR
    		\STATE Sample a minibatch $  x^i, x^i \sim P_{data}(x)$.
    		\STATE Sample  patch indexs $  \alpha_{kl}, \alpha_{kl} \sim P_{0-1}(\alpha)$.
    		\STATE Select  patches of $x^i$ according to $ \alpha_{kl} $.
    		\STATE ${g_{w_G} \leftarrow  \nabla _{w_G} (L _{{w_G}}-\frac{1}{M}\sum\limits_{i = 1}^M {\log (1 - D'(G({y_i})))}}$ 
    		\STATE ${w_G} \leftarrow {w_D} - \eta $ \bf ADAM$(w_G,g_{w_G})$
    		\UNTIL {$w_G$ converges}	
    	\end{algorithmic}
    	\label{algorithm}	
    \end{algorithm}	
	\subsubsection{Complementary Global Structure Regularization}
	The global structure information of the whole image is indispensable for reconstructing CS-MRI to some extent. A new regularization adapted from SSIM is introduced for acquiring the global structure information \cite{wang2004image}.
	
	SSIM is capable of measuring the global structure similarity, so it is applied for pushing neural network to generating images containing similar structure as full-sampled images. Suppose $x'$ and $y'$ are two nonnegative images, $x'$ is full-sampled image that have perfect quality and $ y' $ is generated image with low quality. According to the definition of SSIM, measure satisfy $ SSIM(x',y') \le 1 $, if and only if $ x' = y'$ the equality holds. It is obvious that the value of SSIM increases as the quality of images becomes higher. SSIMR can be defined as follows:
	\begin{equation}
	\ S_{{w_G}}(x',y') = {(1 - SSIM(x',y'))^2}
	\end{equation}
	\subsubsection{Remark 1}
	SSIMR and PCR only constrain solution space of generator not that of the discriminator. And two regularizations capture structure information of CS-MRI images at both local and global scale.

	\subsection{SEGAN algorithm}
Given generator $G$, we sample $N$ real images and $N$ generated images respectively, and solve the maximization problem ${\max _D}\:L_{se}(D,G')$ using ADAM \cite{Kingma2014Adam}. Then given discriminator $D$, we sample a minibatch of real images and randomly select patch index $\alpha_{ij}$, and solve the minimization problem ${\min _G}\:L_{se}(D',G)$. The implementation detail is described in Algorithm \ref{algorithm}.	
\subsection{SU-Net}
For better reconstructed MRI image, generator must have a remarkable capacity for feature extraction. The most effective strategy for feature extraction is the shortcut connection which transfers feature representations from low layers to the high layers\cite{DBLP:conf/cvpr/HeZRS16,huang2017densely}. Here, what we want to emphasize is one type of generator called U-Net
 \cite{DBLP:conf/miccai/2015-3} , that utilizes shortcut connection to fully exploit patch information. Hence, we take U-Net as prototype generator and adapt the model for improving the performance of CS-MRI.

we attempt to strengthen the ability of restoring structure information by means of increasing the number of convolutional filters with different sizes. Multi-scale generator was proposed in Figure \ref{u-net} which has various sizes of convolutional filters in each layer that lays the foundation for optimizing Structure-Enhanced Loss.
\subsubsection{Remark 2}
Another advantage of multi-scale generator is that various filters have multiple scale receptive fields, which benefits the fusion of multi-scale contextual information.

	\subsection{Theoretical Result}
	In this section, we give some theoretical analysis about the Stochastic Structure Enhancing Loss (SSEL) for Compressed Sensing MRI Reconstruction. 
	
	We analysis the convergence of SSEL with Gradient Descent (GD) using the online programming framework proposed in \cite{zinkevich2003online,Kingma2014Adam} . Research of Hinton \cite{tieleman2012lecture,goodfellow2016deep} shows that when optimization algorithm is applied to non-convex function to train a neural network, a locally convex bowl is reached eventually after the learning trajectory pass through many different structures. Under this theory, we analysis the convergence of GD-SSEL in locally convex bowl. An optimization problem of Stochastic Structure-Enhanced Loss consists of a feasible set  $w\in$ $\mathbb{R}^{n}$  and an infinite sequence$ \{ L_{{w_G}}^1,L_{{w_G}}^2,...\}  $ where each $ L_{{w_G}}^t:W \to \mathbb{R} $ is a convex function in a locally convex bowl. 
	At each time $ t$, our goal is to estimation $ w_G $ on a previously unknown function $ T_{{w_G}}^t $. Since it is impossible to get $ T_{{w_G}}^t $ before $t$, we want to evaluate our algorithm using the regret which calculates sum of difference between $ L_{{w_G}}^t $ and the best fixed point $ L_{{w_G^{*}}}^t $.

	\subsubsection{Definition 1}
	Given a convex optimization problem $ (w,\{ L_{{w_G}}^1,L_{{w_G}}^2,...\}) $, if $ \{ w_{{w_G}}^1, w_{{w_G}}^2,...\} $ are the vectors selected by Gradient Descent, then the cost of GD-SSEL until time $ T $ is 
	\begin{equation}
	\ C(T) = \sum\limits_{i = 1}^T {L_{{w_G}}^t}
	\end{equation}
	The best fixed stastic cost until $ T $ is 
	\begin{equation}
	\ C{(T)^*} = {\min _{{w_G} \in W}}C(T) = \sum\limits_{i = 1}^T {L_{w_G^*}^t} 
	\end{equation}
	The regret of GD-SSEL until $T $ is 
	\begin{equation}	
		\ {R_{GD - SSEL}}(T) = C(T) - {C^*}(T)
	\end{equation}

	The goal is to prove that the average regret approaches zero. We show GD-SSEL has $O(\sqrt T )$ regret and proof is given in the supplementary material. We also use some definitions simplify our notation, where $ G_{ij} $ as the output of generator of $ i^{th} $ rows and $ j^{th}$ columns and $ x_{ij} $ as pixel of MRI image of  $ i^{th} $ rows and  $ j^{th} $ columns. Assume that the number of rows  equals to columns is $ K $. We define $ \nabla s \triangleq \frac{{\partial SSIM(x,y)}}{{\partial {w_G}}},\nabla f \triangleq \frac{{\partial f(G(y')_i,G(y')_j)}}{{\partial {w_G}}},\nabla g \triangleq \frac{{\partial {G_{ij}}}}{{\partial {w_G}}} $  and  $ \overline \alpha  {\rm\:{ \triangleq }}\:\mathbb{E}\:{\alpha _{ij}}$.

	\subsubsection{Theorem 1}
	Assume that the function $ f(G(y')_i,G(y')_j) $ is bounded and has bounded gradients, $\lvert f(G(y')_i,G(y')_j)\rvert \le M $ , $ \lVert\nabla f\rVert_2 \le F $ , $ \lVert\nabla g \rVert_2 \le G $ , $ \lVert\nabla s\rVert_2 \le S $ for all $ {w_G}\in \mathbb{R }^{n}  $, distance between any ${{w_G}}^t $ generative by GD-SSEL is bounded, $ \lVert w_G^n-w_G^m \rVert_2\le D$ for any $ m,n \in \{1,...,T\} $ and pixel $ x_{ij}  $ is bounded, $ \lvert x_{ij} - x_{i'j'}\rvert \le d $ for any  $ i,j,i',j' \in \{1,...,K\}$. Let the learning rate $ \eta _t = \dfrac{1}{\sqrt{t}} $ and number of patches is $N$. GD-SSEL achieves the following guarantee, for $T \ge 1$.
	\begin{equation}
	\begin{split}
	\ {R_{GD - SSEL}}(T)& \le \frac{{{D^2}\sqrt T }}{2} + ({\lambda _1}M\overline \alpha  {N^2}F \\ &+ {\lambda _2}S+  {\lambda _3}d{K^2}G)^2(4\sqrt T  - 2)
	\end{split}
	\end{equation}
	
	\subsubsection{Proof Sketch}
	To prove the bound, we expand $ L_{w_G}^t $ in Taylor's series, and prove $ L_{w_G}^t - L_{w_G^*}^t \le (w_G^t-{w_G^t}^*)\nabla _{{w_D}} L_{{w_G}}^t $.
	Then we use skill of inequation zoom for Eq.9 and prove $ {R_{GD - SSEL}}(T) \le \frac{{D^2}\sqrt T }{2} +\frac{{{{\left\| {{\nabla _{{w_G}}}L_{{w_G}}^t} \right\|}^2}}}{2}\sum\limits_{t = 1}^T {{\eta _t}}$. Finally, we decompose the $ {{{{\left\| {{\nabla _{{w_G}}}L_{{w_G}}^t} \right\|}^2}}} $ into three parts including derivative of SPCR , derivative of SSIM and derivative of MSE. 
	
	The main technique for proving this theorem is to utilize the norm inequation zoom and the Taylor's series, then we can prove this theorem.\hfill $\square$
	\subsubsection{Remark 3}
	 Theorem 1 implies that when the function  $ f(G(y')_i,G(y')_j) $  is bounded and has bounded gradients, the summation term is related to number of patches and expectation of random variable $ \alpha $.
	The regret bound suggests reducing the number of patches can improve convergence. Besides form Theorem 1, we can see that the bound gets worse when $ T $ is large. The bound will be reduced to the standard convergence rate $O(\sqrt T )$ as $\lambda _1, \lambda _2,\lambda _3 \to 0 $.
	In comparison to structure enhancing loss, stochastic selected patches to training save time of accessing memory which is suitable for dealing with  big size of images.	

	\subsubsection{Corollary2}
	Assume that the function $ f(G(y')_i,G(y')_j) $ is bounded and has bounded gradients, $\lvert f(G(y')_i,G(y')_j)\rvert \le M $ , $ \lVert\nabla f\rVert_2 \le F $ , $ \lVert\nabla g \rVert_2 \le G $ , $ \lVert\nabla s\rVert_2 \le S $ for all $ {w_G}\in \mathbb{R }^{n}  $, distance between any ${{w_G}}^t $ generative by GD-SSEL is bounded, $ \lVert w_G^n-w_G^m \rVert_2\le D$ for any $ m,n \in \{1,...,T\} $ and pixel $ x_{ij}  $ is bounded, $ \lvert x_{ij} - x_{i'j'}\rvert \le d $ for any  $ i,j,i',j' \in \{1,...,K\}$. Let the learning rate $ \eta _t = \dfrac{1}{\sqrt{t}} $ and number of patches is $N$. GD-SSEL achieves the following guarantee, for $T \ge 1$.
	\begin{equation}
	\mathop {\lim }\limits_{T \to \infty } \frac{{{R_{GD - SSEL}}(T)}}{T} = 0
	\end{equation}
	\emph{Proof}. With Theorem 1, left side of Eq.(11) is thus rewritten as,
	\begin{equation}
	\begin{split}
	\ 0& \le \mathop {\lim }\limits_{T \to \infty } \frac{{{R_{GD - SSEL}}(T)}}{T} \\ &\le \mathop {\lim }\limits_{T \to \infty } \frac{{{D^2}}}{{2\sqrt T }} +  ({\lambda _1}M\overline \alpha  {N^2}F \\ &+ {\lambda _2}S+  {\lambda _3}d{K^2}G)^2(\frac{4}{{\sqrt T }} - \frac{2}{{ T }}) = 0
	\end{split}
	\end{equation} 
	Evidently, Eq.(11) is established.\hfill $\square$
	\subsubsection{Remark 4}
	From the Corollary 2, we can see that the average regret of GD-SSEL approaches zero when $ T\to \infty $. With the increase of iterations GD-SSEL will converge gradually.

	\begin{table*}[t]
	\centering
	
	\caption{ Quantitive reconstruction performance on under-sampling MRI. Experiments are conducted on MRI with different sampling rates, including $10\%$, $20\%$, $30\%$, $40\%$. The average performance as well as standard deviation of each testing set is presented. Compared with other methods, if the performance of SEGAN is
		best among four methods, the corresponding entries are then bolded.}
	\centering  
	\subtable[Sample Rate: 10\% and 20\%]{
		\resizebox{0.98\textwidth}{0.061\textheight}{
			
			\begin{tabular}{c ccccc cc ccccc }
				
				\toprule
				\multirow{2}*{Method} &  \multicolumn{5}{c} {$10\%$}& & &\multicolumn{5}{c} {$20\%$}  \\  % 首列占两行
				\cmidrule{2-6} \cmidrule{9-13}
				&NMSE&SSIM&PSNR&20-LSSM&40-LSSM & & &NMSE&SSIM&PSNR&20-LSSM&40-LSSM \\ \midrule
				Zero-Fill & 0.33$\pm$0.05 &0.68$\pm$0.05&25.07$\pm$3.52&0.75$\pm$0.05 &0.75$\pm$0.06 && &0.19$\pm$0.04&0.85$\pm$0.02&29.53$\pm$4.25&0.88$\pm$0.03&0.88$\pm$0.03 \\ 
				
				ADMM-Net & 0.24$\pm$0.04 &0.70$\pm$0.04&30.70$\pm$3.54&0.78$\pm$0.06&0.79$\pm$0.06 && &0.16$\pm$0.03&0.88$\pm$0.04&37.31$\pm$5.25&0.93$\pm$0.01&0.92$\pm$0.02 \\ 
				
				ReGAN & \bf{0.17$\pm$0.04} &0.83$\pm$0.02&33.54$\pm$3.56&0.83$\pm$0.05&0.82$\pm$0.05 & & &0.09$\pm$0.03&0.92$\pm$0.04&39.02$\pm$5.24&0.93$\pm$0.02&0.92$\pm$0.02 \\ 
				
				DAGAN & 0.19$\pm$0.03 &0.82$\pm$0.04&32.84$\pm$2.56&0.82$\pm$0.04&0.81$\pm$0.04 && &0.10$\pm$0.02&0.95$\pm$0.03&38.88$\pm$3.26&0.93$\pm$0.02&0.92$\pm$0.02 \\ 
				
				SEGAN & \bf{0.17$\pm$0.04} &\bf{0.84$\pm$0.01}&\bf{34.60$\pm$3.52}&\bf{0.84$\pm$0.03}&\bf{0.83$\pm$0.03} & & &\bf{0.08$\pm$0.01}&\bf{0.96$\pm$0.01}&\bf{40.80$\pm$2.68}&\bf{0.96$\pm$0.01}&\bf{0.95$\pm$0.01} \\ 
				\bottomrule
				
			\end{tabular}
		}
	}
	\centering  
	\subtable[Sample Rate: 30\% and 40\%]{
		\resizebox{0.98\textwidth}{0.061\textheight}{
			
			\begin{tabular}{c ccccc cc ccccc } 
				
				\toprule
				\multirow{2}*{Method} &  \multicolumn{5}{c} {$30\%$}& & &\multicolumn{5}{c} {$40\%$}  \\  % 首列占两行
				\cmidrule{2-6} \cmidrule{9-13}
				&NMSE&SSIM&PSNR&20-LSSM&40-LSSM & & &NMSE&SSIM&PSNR&20-LSSM&40-LSSM \\ \midrule
				Zero-Fill & 0.18$\pm$0.02 &0.86$\pm$0.01&30.80$\pm$3.40&0.90$\pm$0.03& 0.90$\pm$0.03& &  &0.15$\pm$0.03&0.90$\pm$0.02&32.63$\pm$4.50&0.92$\pm$0.02&0.92$\pm$0.02 \\ 
				ADMM-Net & 0.15$\pm$0.01 &0.91$\pm$0.02&37.36$\pm$4.55&0.94$\pm$0.03&0.93$\pm$0.03 && &0.10$\pm$0.02&0.95$\pm$0.01&42.36$\pm$3.62&0.96$\pm$0.02&0.96$\pm$0.02 \\ 
				ReGAN & 0.09$\pm$0.01 &0.96$\pm$0.02&39.93$\pm$3.68&0.95$\pm$0.01&\bf{0.96$\pm$0.01} & & &0.06$\pm$0.02&\bf{0.98$\pm$0.01}&43.43$\pm$3.75&0.97$\pm$0.01&\bf{0.97$\pm$0.01} \\ 
				DAGAN & 0.09$\pm$0.02 &0.96$\pm$0.01&39.53$\pm$4.08&0.94$\pm$0.02&0.93$\pm$0.02 && &0.06$\pm$0.01&0.96$\pm$0.02&43.28$\pm$3.62&0.95$\pm$0.01&0.95$\pm$0.01 \\ 
				SEGAN & \bf{0.07$\pm$0.01} &\bf{0.97$\pm$0.01}&\bf{41.50$\pm$4.52}&\bf0.97$\pm$0.01&\bf0.96$\pm$0.01 & & &\bf{0.05$\pm$0.01}&\bf{0.98$\pm$0.01}&\bf{44.15$\pm$3.55}&\bf{0.98$\pm$0.01}&\bf{0.97$\pm$0.01} \\ 
				\bottomrule
			\end{tabular}
		}
	}
	\label{tab_re}
\end{table*}

	\section{Experiments}
	The previous section describes our method SEGAN and gives some theoretical analysis. In this section we compare our algorithm with several state-of-art methods on two aspects:
\begin{enumerate}[label=\arabic*)]
	\item Reconstruction Performance: we will examine the performance of SEGAN in CS-MRI reconstruction.
	\item Testing Time: we will examine the time of generating an image in CS-MRI reconstruction.
\end{enumerate}	

	\subsubsection{Dataset }
We use a MICCAI 2013 grand challenge dataset and randomly split T1 weight MRI dataset, 16095 for training, 5033 for validation and 9854 for testing independently. We test our algorithm under different underdampling masks, gaussian masks with 10\%, 20\%, 30\%, 40\% sub-sampling rates, which enables 10$ \times $, 5$ \times $, 3.3$ \times $ and 2.5$ \times $ accelerations. In the experiment, we use zero-filling under-sampled images as input.
	\subsubsection{Implementation Details }
	Training and testing the algorithm use tensorflow with the python environment on a NVIDIA GeFore GTX TITAN X with 12GB GPU memory. We used ADAM with momentum for parameter optimization and set the initial learning rate to be 0.0001, the first-order momentum to be 0.9 and the second momentum to be 0.999. The weight decays regularization parameter is 0.0001 and the batch size is 30. 30000 stochastic iterations of training were required to train the SEGAN. For all experiments we use three scales of convolutional filters that are 2$ \times $2, 4$ \times $4 and 6$ \times $6 in the generator and a standard CNN architecture with 11 convolutional layers with $3 \times 3$ convolutional filters in the discriminator which are also followed by batch normalization and leaky ReLU layers.
	
	We compared our SEGAN model with state-of-art models in the latest development including ADMM-Net \cite{Yang2017ADMM}, ReGAN \cite{Quan2018Compressed}, DAGAN \cite{Yang2017DAGAN}. The performance of other conventional methods is much poorer than that of SEGAN, so we did not show the results. For ADMM-Net method, we performed 600 iterations for obtaining the reconstruction to avoid the overfit. For DAGAN and ReGAN method, we use the software provided by the authors on github and fine-tuned parameter settings according to experiment results. 
	\subsubsection{Training Specifications }
	 In summary, the SSEL of SU-Net involves three portions which correspond different goals for generating images containing rich structure information. $ {\lambda _1},{\lambda _2}$ and ${\lambda _3} $ are the trade-off parameters which help to balance the importance of each contribution. In the experiment, We empirically set $ {\lambda _1} = 10 $, $ {\lambda _2} = 1$ and $ {\lambda _3} = 100 $ to maximize reconstruction performance. We select the polynomial kernel function as correlation function. The number of patches is fixed to 64 and $ \overline \alpha $ is set to 0.1.
	 \begin{figure}[t]
	 	\centering
	 	\includegraphics[width=0.46\textwidth]{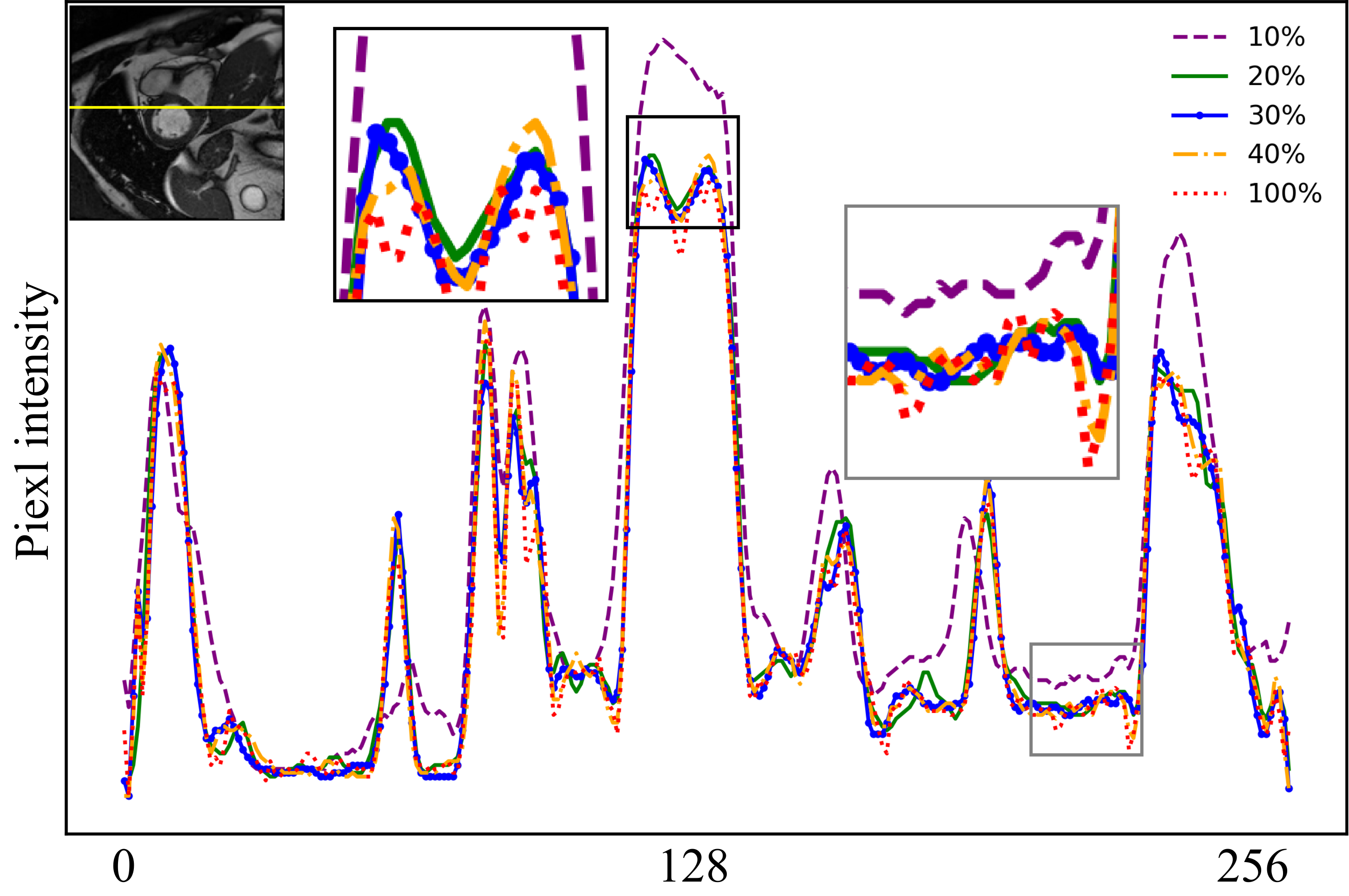}
	 	\centering
	 	\caption{Illustration of image quality comparison on different sample rates. Curves painted in different colors represent images under distinct sample rates. The horizontal coordinate denotes pixel positions of the yellow line drawn in the original image, and the vertical coordinate is pixel intensity of the corresponding pixel. Clearly, the better matching with red curves means superior generation quality.}
	 	\label{maskpresent}
	 \end{figure}

\begin{figure}
	\includegraphics[width=0.46\textwidth]{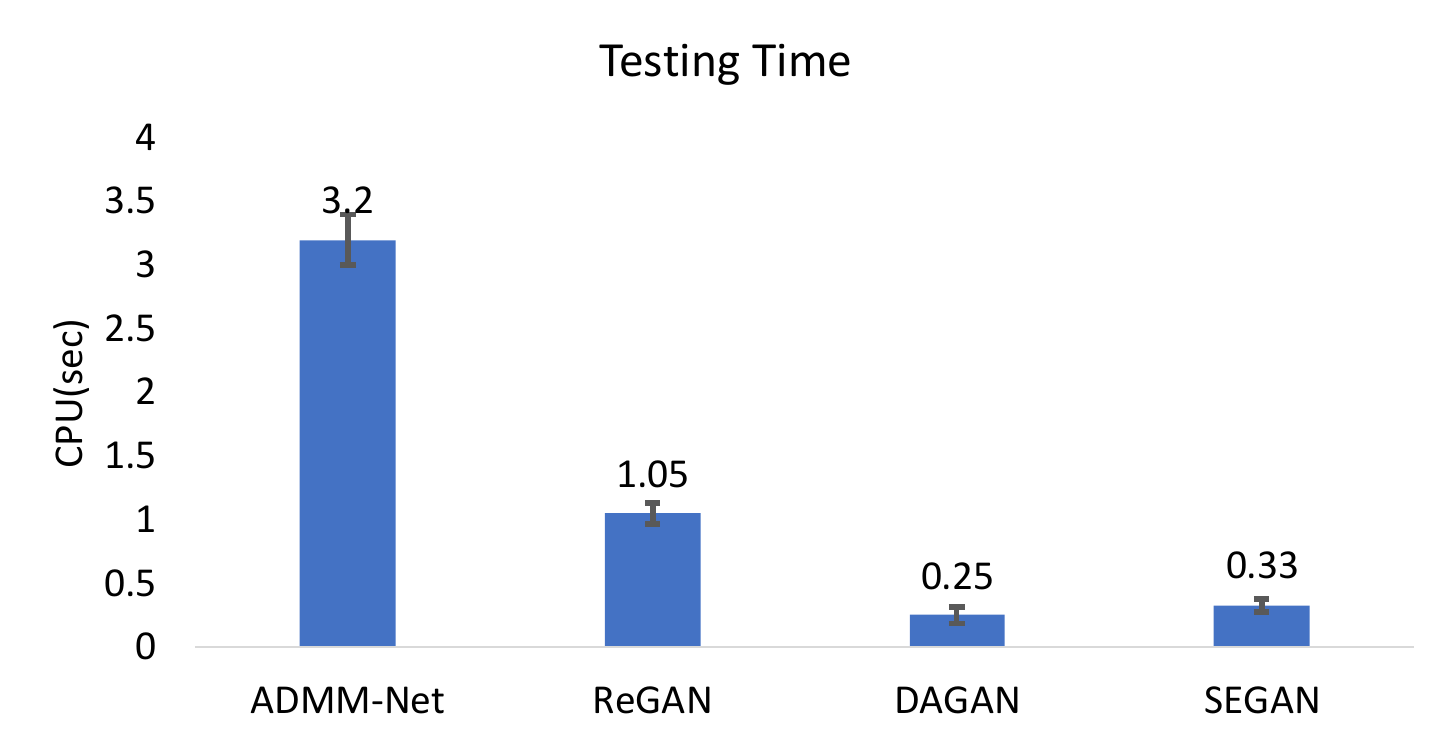}
	\centering	
	\caption{Illustration of testing time of different methods. Shorter testing time makes reconstruction method more possible to be applied in clinical medicine.}
	\centering
	\label{time}	
\end{figure}

	\subsubsection{Evaluation Methods  }
	 We use Normalized Mean Square Error (NMSE), the Peak signal-to-noise Ratio (PSNR) and SSIM to evaluate the performance of reconstructed images. In order to further quantify the restored local correlation, we introduce a  measurement called N patches Local Structure Similary Mean  (N-LSSM) that is the average of patch structure similarly between generated images and ground truth. Concretely, we randomly pick up N patches of reconstructed images and calculate SSIM between patches that are selected from generated image and  ground truth. Obviously, larger N-LSSM indicates better reconstructed quality. In the experiment, we fix the number of patches to 20 and 40, the size of each patch is $20\times20$.
\begin{figure*}[h]
	\centering
	\includegraphics[width=0.99\textwidth]{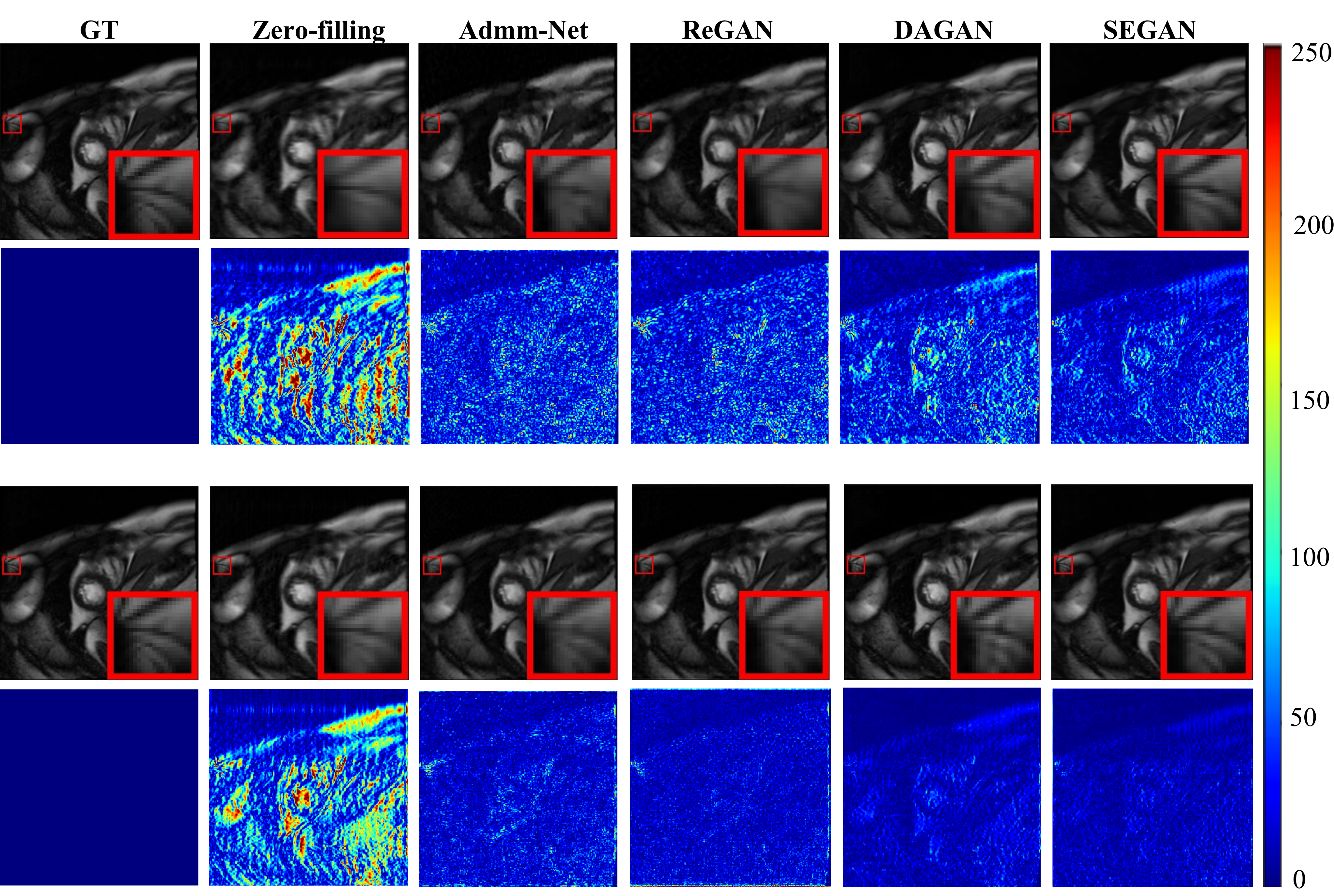}
	\centering
	\caption{Illustration of reconstruction image quality comparison between different methods at the two sample rates including 20\% (top 2 rows) and 40\% (bottom 2 rows). The second and forth row is error maps which zoom the error 10 times between reconstruction and original image. Error map is showed with the help of color lumps, more blue lumps represents more similarity compared with the original image. }
	\label{resultpresent}
\end{figure*}
	\subsection{Reconstruction Performance}
	\subsubsection{Compare with Different Sampling Rate}
	It is observed from numerical results that sampling rate plays the key role in the quality of reconstruction. Specifically Figure \ref{maskpresent} and Table \ref{tab_re} show that the higher sampling rate comes with smaller MSE, larger PSNR and SSIM. Apart from the former three measures, what we want to emphasize is that N-LSSM increases as higher sample rates as well. Note that, there is a significant improvement of performance when sampling rate increases from 10\% to 20\%. Conversely, the performance improvement is not that significant when sample rate increased from 20\% to 40\%.
	
	\subsubsection{Compare with Other Methods}
	As shown in Table \ref{tab_re} and Figure \ref{resultpresent}, the performance of SEGAN has been improved in terms of different criteria with respective to different sampling rates. Additionally, our method SEGAN produced visually reconstructions that outperform the other algorithms. From the perspective of structure information recovery, reconstructed MRI image of SEGAN  is the best both at local and global scale. 

\subsection{Testing Time}
 Considering that computational resources in realistic medical environment are limited, the testing time of reconstruction directly determines the application scope of the method. We test the reconstruction time of several main stream models, specifically, we test those methods under the environment of Intel Xeon CPU E5-1603 with 32GB memory. The testing time is show in Figure \ref{time}. As shown that, the inference time of several deep generative models is shorter than ADMM-Net, the inference time of our method is slightly higher than the minimal value 1.05s of DAGAN, but SEGAN achieves much better performance than other models. 
\section{Conclusion}
In this paper, we propose a novel deep generative model called SEGAN for CS-MRI reconstruction, that enable the restoration of structure information contained in full-sampled MRI images. SEGAN can uncover the local structure correlation consisting in different patches by the proposed structure regularization PCR. To speed up the training and reduce the requirement of memory size, we propose a stochastic PCR in which patches are randomly selected from integral patch partition. In addition, a novel deep architecture of generator called SU-Net is designed for efficiently extract local structure correlation at various scale.

Besides, we theoretically analyze convergence of stochastic PCR under the condition of gradient descent based optimization. Moreover, the superior performance of reconstructed MRI images generated by SEGAN is validated in experimental parts with state-of-the-art methods.
In the future, it is interesting  to propose new structure regularizations that are derived from new pixel correlation measure methods and easier to be optimized under the framework of deep architecture.
%one interesting issue is to explore new network architecture dedicated to MRI image reconstruction. Another

\bibliographystyle{aaai}
\bibliography{refence1}

\begin{thebibliography}{}

\bibitem[\protect\citeauthoryear{Aharon \bgroup et al\mbox.\egroup
  }{2006}]{aharon2006k}
Aharon, M.; Elad, M.; Bruckstein, A.; et~al.
\newblock 2006.
\newblock {K-SVD}: An algorithm for designing overcomplete dictionaries for
  sparse representation.
\newblock {\em IEEE Transactions on signal processing} 54(11):4311.

\bibitem[\protect\citeauthoryear{Caballero \bgroup et al\mbox.\egroup
  }{2014}]{caballero2014dictionary}
Caballero, J.; Price, A.~N.; Rueckert, D.; and Hajnal, J.~V.
\newblock 2014.
\newblock Dictionary learning and time sparsity for dynamic mr data
  reconstruction.
\newblock {\em IEEE Transactions on medical imaging}  979--994.

\bibitem[\protect\citeauthoryear{Cand{\`e}s and
  Wakin}{2008}]{candes2008introduction}
Cand{\`e}s, E.~J., and Wakin, M.~B.
\newblock 2008.
\newblock An introduction to compressive sampling.
\newblock {\em IEEE signal processing magazine} 25(2):21--30.

\bibitem[\protect\citeauthoryear{Cand{\`e}s, Romberg, and
  Tao}{2006}]{candes2006robust}
Cand{\`e}s, E.~J.; Romberg, J.; and Tao, T.
\newblock 2006.
\newblock Robust uncertainty principles: Exact signal reconstruction from
  highly incomplete frequency information.
\newblock {\em IEEE Transactions on information theory}  489--509.

\bibitem[\protect\citeauthoryear{Goldstein and
  Osher}{2009}]{goldstein2009split}
Goldstein, T., and Osher, S.
\newblock 2009.
\newblock The split bregman method for l1-regularized problems.
\newblock {\em SIAM journal on imaging sciences} 2(2):323--343.

\bibitem[\protect\citeauthoryear{Goodfellow \bgroup et al\mbox.\egroup
  }{2014}]{goodfellow2014generative}
Goodfellow, I.; Pouget-Abadie, J.; Mirza, M.; Xu, B.; Warde-Farley, D.; Ozair,
  S.; Courville, A.; and Bengio, Y.
\newblock 2014.
\newblock Generative adversarial nets.
\newblock In {\em Advances in Neural Information Processing Systems(NIPS
  2014)},  2672--2680.

\bibitem[\protect\citeauthoryear{Goodfellow \bgroup et al\mbox.\egroup
  }{2016}]{goodfellow2016deep}
Goodfellow, I.; Bengio, Y.; Courville, A.; and Bengio, Y.
\newblock 2016.
\newblock {\em Deep learning}, volume~1.
\newblock MIT press Cambridge.

\bibitem[\protect\citeauthoryear{Hammernik \bgroup et al\mbox.\egroup
  }{2017}]{HammernikKKRSPK17}
Hammernik, K.; Klatzer, T.; Kobler, E.; Recht, M.~P.; Sodickson, D.~K.; Pock,
  T.; and Knoll, F.
\newblock 2017.
\newblock Learning a variational network for reconstruction of accelerated
  {MRI} data.
\newblock abs/1704.00447.

\bibitem[\protect\citeauthoryear{He \bgroup et al\mbox.\egroup
  }{2016}]{DBLP:conf/cvpr/HeZRS16}
He, K.; Zhang, X.; Ren, S.; and Sun, J.
\newblock 2016.
\newblock Deep residual learning for image recognition.
\newblock In {\em Proceedings of the IEEE conference on computer vision and
  pattern recognition(CVPR 2016)},  770--778.

\bibitem[\protect\citeauthoryear{Hollingsworth}{2015}]{hollingsworth2015reducing}
Hollingsworth, K.~G.
\newblock 2015.
\newblock Reducing acquisition time in clinical {MRI} by data undersampling and
  compressed sensing reconstruction.
\newblock {\em Physics in Medicine \& Biology}  R297.

\bibitem[\protect\citeauthoryear{Huang \bgroup et al\mbox.\egroup
  }{2017}]{huang2017densely}
Huang, G.; Liu, Z.; Van Der~Maaten, L.; and Weinberger, K.~Q.
\newblock 2017.
\newblock Densely connected convolutional networks.
\newblock In {\em Proceedings of the IEEE conference on computer vision and
  pattern recognition(CVPR 2017)},  3--11.

\bibitem[\protect\citeauthoryear{Jia, Lu, and Yang}{2012}]{jia2012visual}
Jia, X.; Lu, H.; and Yang, M.-H.
\newblock 2012.
\newblock Visual tracking via adaptive structural local sparse appearance
  model.
\newblock In {\em Proceedings of the IEEE conference on computer vision and
  pattern recognition(CVPR 2012)},  1822--1829.

\bibitem[\protect\citeauthoryear{Kingma and Ba}{2014}]{Kingma2014Adam}
Kingma, D.~P., and Ba, J.
\newblock 2014.
\newblock Adam:{A} method for stochastic optimization.
\newblock In {\em International Conference on Learning Representations(ICML
  2014)}.

\bibitem[\protect\citeauthoryear{Lai \bgroup et al\mbox.\egroup
  }{2016}]{lai2016image}
Lai, Z.; Qu, X.; Liu, Y.; Guo, D.; Ye, J.; Zhan, Z.; and Chen, Z.
\newblock 2016.
\newblock Image reconstruction of compressed sensing mri using graph-based
  redundant wavelet transform.
\newblock {\em Medical image analysis}  93--104.

\bibitem[\protect\citeauthoryear{Lee, Yoo, and Ye}{2017}]{Lee2017Deep}
Lee, D.; Yoo, J.; and Ye, J.~C.
\newblock 2017.
\newblock Deep residual learning for compressed sensing {MRI}.
\newblock In {\em 2017 IEEE 14th International Symposium on Biomedical Imaging
  (ISBI 2017)},  15--18.

\bibitem[\protect\citeauthoryear{Lustig, Donoho, and
  Pauly}{2007}]{lustig2007sparse}
Lustig, M.; Donoho, D.; and Pauly, J.~M.
\newblock 2007.
\newblock Sparse {MRI}: The application of compressed sensing for rapid {MR}
  imaging.
\newblock {\em Magnetic Resonance in Medicine}  1182--1195.

\bibitem[\protect\citeauthoryear{Navab \bgroup et al\mbox.\egroup
  }{2015}]{DBLP:conf/miccai/2015-3}
Navab, N.; Hornegger, J.; III, W. M.~W.; and Frangi, A.~F.
\newblock 2015.
\newblock {U-Net}: Convolutional networks for biomedical image segmentation.
\newblock In {\em Medical Image Computing and Computer-Assisted
  Intervention(MICCAI 2015)},  234--241.

\bibitem[\protect\citeauthoryear{Otazo, Cand{\`e}s, and
  Sodickson}{2015}]{otazo2015low}
Otazo, R.; Cand{\`e}s, E.; and Sodickson, D.~K.
\newblock 2015.
\newblock Low-rank plus sparse matrix decomposition for accelerated dynamic mri
  with separation of background and dynamic components.
\newblock {\em Magnetic Resonance in Medicine}  1125--1136.

\bibitem[\protect\citeauthoryear{Quan, Nguyenduc, and
  Jeong}{2018}]{Quan2018Compressed}
Quan, T.~M.; Nguyenduc, T.; and Jeong, W.~K.
\newblock 2018.
\newblock Compressed sensing {MRI} reconstruction using a generative
  adversarial network with a cyclic loss.
\newblock {\em IEEE Transactions on medical imaging} 37(6):1488--1497.

\bibitem[\protect\citeauthoryear{Schlemper \bgroup et al\mbox.\egroup
  }{2017}]{SchlemperCHPR17}
Schlemper, J.; Caballero, J.; Hajnal, J.~V.; Price, A.~N.; and Rueckert, D.
\newblock 2017.
\newblock A deep cascade of convolutional neural networks for {MR} image
  reconstruction.
\newblock In {\em Information Processing in Medical Imaging},  647--658.

\bibitem[\protect\citeauthoryear{Shen, Wu, and Suk}{2017}]{shen2017deep}
Shen, D.; Wu, G.; and Suk, H.-I.
\newblock 2017.
\newblock Deep learning in medical image analysis.
\newblock {\em Annual review of biomedical engineering}  221--248.

\bibitem[\protect\citeauthoryear{Tieleman and
  Hinton}{2012}]{tieleman2012lecture}
Tieleman, T., and Hinton, G.
\newblock 2012.
\newblock Lecture 6.5-{RMSProp, COURSERA}: Neural networks for machine
  learning.
\newblock {\em University of Toronto, Technical Report}.

\bibitem[\protect\citeauthoryear{Tr{\'e}moulh{\'e}ac \bgroup et al\mbox.\egroup
  }{2014}]{tremoulheac2014dynamic}
Tr{\'e}moulh{\'e}ac, B.; Dikaios, N.; Atkinson, D.; and Arridge, S.~R.
\newblock 2014.
\newblock Dynamic mr image reconstruction--separation from undersampled
  {(K,t)}-space via low-rank plus sparse prior.
\newblock {\em IEEE Transactions on medical imaging}  1689--1701.

\bibitem[\protect\citeauthoryear{Wang \bgroup et al\mbox.\egroup
  }{2004}]{wang2004image}
Wang, Z.; Bovik, A.~C.; Sheikh, H.~R.; and Simoncelli, E.~P.
\newblock 2004.
\newblock Image quality assessment: from error visibility to structural
  similarity.
\newblock {\em IEEE transactions on image processing} 13(4):600--612.

\bibitem[\protect\citeauthoryear{Yang \bgroup et al\mbox.\egroup
  }{2017a}]{Yang2017DAGAN}
Yang, G.; Yu, S.; Dong, H.; Slabaugh, G.; Dragotti, P.~L.; Ye, X.; Liu, F.;
  Arridge, S.; Keegan, J.; and Guo, Y.
\newblock 2017a.
\newblock {DAGAN}: Deep de-aliasing generative adversarial networks for fast
  compressed sensing {MRI} reconstruction.
\newblock {\em IEEE Transactions on medical imaging} PP(99):1--1.

\bibitem[\protect\citeauthoryear{Yang \bgroup et al\mbox.\egroup
  }{2017b}]{Yang2017ADMM}
Yang, Y.; Sun, J.; Li, H.; and Xu, Z.
\newblock 2017b.
\newblock {ADMM-Net}: A deep learning approach for compressive sensing {MRI}.
\newblock In {\em Advances in Neural Information Processing Systems(NIPS
  2017)},  10--18.

\bibitem[\protect\citeauthoryear{Yao \bgroup et al\mbox.\egroup
  }{2015}]{yao2015accelerated}
Yao, J.; Xu, Z.; Huang, X.; and Huang, J.
\newblock 2015.
\newblock Accelerated dynamic {MRI} reconstruction with total variation and
  nuclear norm regularization.
\newblock In {\em International Conference on Medical Image Computing and
  Computer-Assisted Intervention(MICCAI 2015)},  635--642.

\bibitem[\protect\citeauthoryear{Zhan \bgroup et al\mbox.\egroup
  }{2016}]{zhan2016fast}
Zhan, Z.; Cai, J.-F.; Guo, D.; Liu, Y.; Chen, Z.; and Qu, X.
\newblock 2016.
\newblock Fast multiclass dictionaries learning with geometrical directions in
  mri reconstruction.
\newblock {\em IEEE Transactions on biomedical engineering}  1850--1861.

\bibitem[\protect\citeauthoryear{Zinkevich}{2003}]{zinkevich2003online}
Zinkevich, M.
\newblock 2003.
\newblock Online convex programming and generalized infinitesimal gradient
  ascent.
\newblock In {\em Proceedings of the 20th International Conference on Machine
  Learning(ICML 2003)},  928--936.

\end{thebibliography}

\end{document}